\documentclass[acmtog, screen, nonacm]{acmart}
\AtBeginDocument{%
  }

\usepackage{amsfonts}
\usepackage{amsmath}
\usepackage{amsthm}
\usepackage{bm}
\usepackage{multirow}
\usepackage{graphicx}
\usepackage{enumitem}
\usepackage{array}
\usepackage{wrapfig}
\usepackage{xcolor}
\usepackage{algorithm}
\usepackage{nicefrac}
\usepackage{microtype}
\usepackage{pifont}
\usepackage{subcaption}

\newcommand{\colref}[3]{\hyperref[#2]{#1~\ref*{#2}{#3}}}
\newcommand{\figref}[1]{\colref{Figure}{#1}{}}

\newcommand{\secref}[1]{\colref{Section}{#1}{}}

\newcommand{\tabref}[1]{\colref{Table}{#1}{}}

\begin{document}

%%
%% The "title" command has an optional parameter,
%% allowing the author to define a "short title" to be used in page headers.
\title{Linkify: Learning from Interface-Augmented Assembly Graphs}

%%
%% The "author" command and its associated commands are used to define
%% the authors and their affiliations.
%% Of note is the shared affiliation of the first two authors, and the
%% "authornote" and "authornotemark" commands
%% used to denote shared contribution to the research.
\author{Anushrut Jignasu}
\authornote{Corresponding author. Work done while interning at Autodesk.}
% \correspondingauthor
% \authornotemark[1]
\email{ajignasu@iastate.edu}
\affiliation{%
  \institution{Iowa State University}
  \country{USA}
}

\author{Daniele Grandi}
\email{daniele.grandi@autodesk.com}
\affiliation{%
  \institution{Autodesk Research}
  \country{USA}}

%%
%% By default, the full list of authors will be used in the page
%% headers. Often, this list is too long, and will overlap
%% other information printed in the page headers. This command allows
%% the author to define a more concise list
%% of authors' names for this purpose.
\renewcommand{\shortauthors}{Anushrut Jignasu and Daniele Grandi}

%%
%% The abstract is a short summary of the work to be presented in the
%% article.
\begin{abstract}
  We present Linkify, a framework for learning from interface-augmented assembly graphs to enable context-aware part retrieval in mechanical assemblies. While recent generative AI methods for CAD have focused largely on isolated parts or monolithic assemblies, the rich geometric information at the interfaces between parts, where function is realized, remains underexplored. We address this gap by recomputing high-fidelity interface geometry for the Fusion 360 Gallery Assembly dataset, correcting missing and erroneous contacts, and generating point-cloud representations of local contact regions. Using this data, we construct assembly graphs whose nodes encode part geometry and whose edges encode interface geometry via a pretrained point-cloud encoder. On top of this representation, we train a Graph Attention Network based on GATv2 to solve a masked part prediction task: given an assembly with one part held out, the model predicts the class of the missing component from a large vocabulary of geometrically clustered parts, thereby approximating a realistic part-retrieval scenario. Compared to non-graph baselines such as logistic regression and k-nearest neighbors operating on aggregated node features, Linkify achieves higher Top-K accuracy and F1 scores. Ablation studies on graph connectivity, edge attributes, and attention mechanisms demonstrate that accurate contact computation and dynamic attention over interfaces are critical for performance. Our corrected interface dataset and training pipeline, released publicly, provide a foundation for future interface-aware models for assembly retrieval, validation, and generative design.
\end{abstract}

\begin{teaserfigure}
  \centering
  \includegraphics[width=0.99\linewidth] {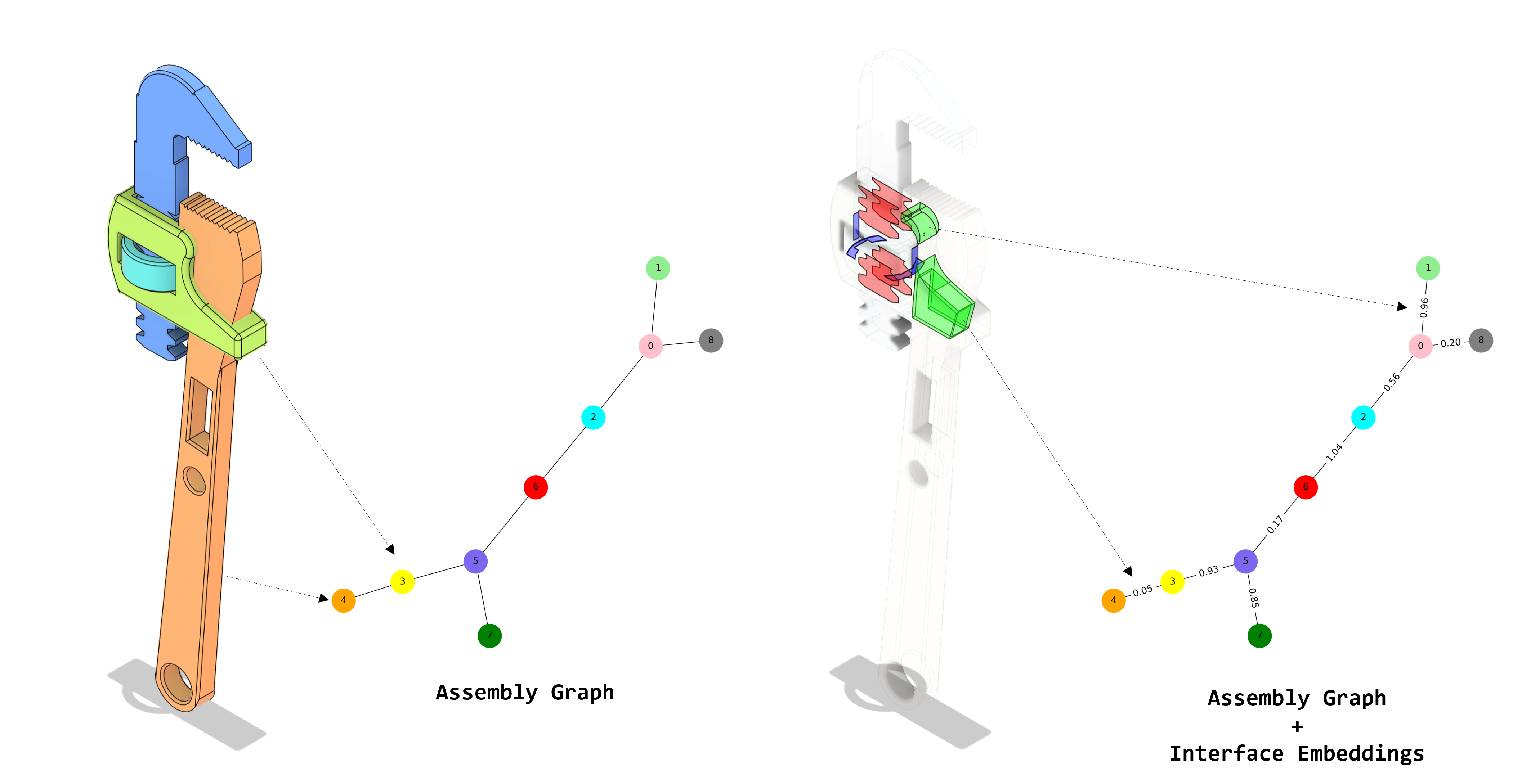}
  \Description{A horizontal teaser image showing the assembly graph representation with embedded local interface information.}
	\captionof{figure}{We augment the assembly graph representation, where part geometries are embedded within nodes of the graph, by embedding local interface information (contact geometry) within the edges of the graph.}
  \label{fig:teaser}
\end{teaserfigure}

%%
%% This command processes the author and affiliation and title
%% information and builds the first part of the formatted document.
\maketitle

\section{Introduction}
\label{intro}

In recent years, the field of computer-aided design (CAD) has seen a surge of interest in leveraging generative AI to automate and accelerate the detailed design process. Much of the work has concentrated on generating single, isolated parts or monolithic fused assemblies without moving parts. The more complex challenge of designing multi-component assemblies remains a frontier. Current approaches have been trained on assembly graphs where nodes represent individual parts, and edges represent kinematic joints or, seldom, hierarchical part structures defined by CAD users. However, the rich and functional information embedded in the precise interface geometry (or contacts) between interacting parts remains understudied. In this work, we address this gap by exploring how interfaces can be represented and how they enable machine learning (ML) models to learn from this critical data, which defines how an assembly of parts fits together and functions. To this end, we propose an augmented assembly graph data structure that explicitly utilizes these geometric interface features (\figref{fig:teaser}), intended as an initial step toward leveraging interfaces for downstream learning tasks.

Although this enhanced data structure is a key enabler of generative AI for assembly design, in this paper, we focus on a more fundamental engineering challenge: part retrieval. In large engineering enterprises, engineers and designers often spend significant time recreating parts that are functionally identical or similar to existing, validated components already in the company's database. This design redundancy increases costs, complicates supply chains, and extends development timelines. A potential solution to this is an intelligent part retrieval system that can understand a part's functional context.

Effective, context-aware part retrieval offers three primary benefits. First, it can act as design autocomplete, proactively suggesting appropriate previously designed parts based on the current assembly context, thereby reducing the time engineers spend searching through vast component libraries. Second, it serves as a powerful design validation tool by checking whether a newly chosen or designed part is consistent with historical usage patterns for similar applications. Finally, by analyzing the retrieval patterns and feature importance, we can gain insights into design intent, learning which geometric or material properties are most critical for a given function and feeding this knowledge back to designers to improve future creations.

Traditionally, part retrieval has relied on techniques like text-based metadata search or shape similarity matching of whole parts. While useful, these methods have a critical limitation: they are context-agnostic. They do not answer the question ``What part fits here?'' because they do not consider how a component interacts with its neighbors within a specific assembly. This is primarily because they lack a formal representation of the geometric and functional constraints imposed by the assembly environment.

In this paper, we introduce an interface-aware dataset, and as a baseline, we introduce Linkify, a framework that leverages interface-aware assembly graphs to learn rich, context-dependent representations of mechanical parts. To validate our approach and test the central hypothesis that interface geometry is critical for understanding a part's function, we tackle part retrieval formulated as a masked part prediction task. Given an assembly graph with one component held out, the model must predict its identity from a large vocabulary of known parts. Solving this multi-class classification problem serves as a proxy for a much more complex assembly autocomplete scenario, where, along with interfaces, we would also predict the interface axis, optimal area of contact, and its associated functional nature. We emphasize that this formulation is intended as a retrieval-oriented probe task, not as a complete assembly synthesis or fit-verification system. In particular, the predicted output is a cluster label for the missing part rather than a direct guarantee that a retrieved candidate is geometrically, kinematically, or functionally assembleable with the surrounding components.

Our primary contributions are as follows:
\vspace{-0.3em}
\begin{enumerate}
    \item We augment the Fusion 360 Gallery Assembly dataset with high-fidelity geometric interface definitions, creating a new resource for building and exploring interface-aware assembly graphs.
    \item We build a Graph Neural Network (GNN) architecture based on \texttt{GATv2}, demonstrating how interface-augmented assembly graphs can be used within a learning pipeline.
    \item We conduct a series of ablation studies to quantitatively measure the feature importance of graph topology and interface geometry, confirming that our model successfully leverages this information.
\end{enumerate}

More specifically, our experiments are guided by the following research questions (RQ): 

\begin{enumerate}
    \item How critical is the correct computation of the edges representing contacts when learning from assembly graphs?
    \item Does embedding interface geometry as edge attributes in the assembly graphs help the model learn a better assembly representation?
    \item Does the dynamic attention mechanism of GATv2 offer an advantage over the standard GAT?
\end{enumerate}

To support further development of models that learn from interface data in assembly graphs and to evaluate our method, the corrected interface information for the Fusion 360 Gallery dataset and our training pipeline are available on Github\footnote{\href{https://github.com/ajignasu/linkify}{https://github.com/ajignasu/linkify}}.

\section{Related Work}\label{sec:relatedwork}

\subsection{Graph Neural Networks}
Graph Neural Networks (GNNs) are a class of deep neural networks designed for learning and inference on graph-structured data distributed in non-Euclidean spaces~\cite{zhang_2020_kde, wu_2020_nnls}. Representation learning in GNNs relies on message passing, which typically involves two key steps: aggregation, where features are collected from neighboring nodes and edges, and combination, where the aggregated features are integrated into an updated node representation. This iterative process allows GNNs to capture order-invariant and variable-sized graph structures with topological awareness \cite{li_2015_iclr, gilmer_2017_icml, kipf_2017_iclr, velickovic_2018_iclr, xu_2019_iclr, Khasahmadi2020Memory}. The resulting node-level representations are further pooled to generate graph-level representations, enabling tasks such as node, edge, and graph classification.

Beyond generic graph domains, GNN-inspired architectures have demonstrated strong representational power for learning geometric relationships. DGCNN~\cite{wang2019dynamic} constructs a dynamic k-nearest neighbor graph in feature space and applies the EdgeConv operator, a localized aggregation over edge features, to capture local geometric structure within point clouds. More recent frameworks like ROCA~\cite{gumeli2022roca} and DiffCAD~\cite{gao2024diffcad} extend this idea to CAD model retrieval and alignment, where embeddings learned from 3D geometry preserve topological and shape similarity.

In the broader engineering domain, GNNs have been applied to physics-informed and design-aware learning tasks. In computational fluid dynamics (CFD), MeshGraphNets~\cite{pfaff2020learning} treat mesh vertices as graph nodes and learn to propagate physical quantities through message passing, enabling differentiable surrogate models for PDE solvers. Similar ideas appear in materials modeling~\cite{dai2021graph}, where microstructural or atomic systems are represented as interaction graphs to predict mechanical or functional properties.

Within CAD and design automation, GNNs have become increasingly prominent for representing geometric and relational information in engineering models. In parametric CAD, sketches and constraints can be represented as graphs where nodes denote geometric primitives and edges encode relationships such as parallelism or coincidence~\cite{seff2020sketchgraphs}. In boundary representation (B-rep) CAD models, nodes correspond to topological entities (faces, edges, vertices) and edges capture adjacency relationships, enabling model classification and retrieval directly from native CAD structures without conversion to meshes~\cite{mandelli2022cad}. At the assembly level, GNNs naturally model inter-part relationships: JoinABLe~\cite{willis2022joinable} predicts joint alignments between parts, while~\cite{gajek2022recommendation} uses graph attention networks to recommend missing components in partial assemblies. More recently, HG-CAD~\cite{bian2024hg} integrated part-level geometry and assembly-level structure within a hierarchical graph to predict material assignments for CAD components.

The choice of aggregation and combination functions is critical, as they give rise to different variants of GNNs tailored for specific tasks. In the standard Graph Convolutional Network (GCN) \cite{kipf_2017_iclr}, the aggregation function averages the features of neighboring nodes, while the combination function applies a linear transformation followed by a nonlinearity. Variations arise when these functions are modified or made more expressive. For example, in the Graph Attention Network (GAT) \cite{velickovic_2018_iclr}, the aggregation function employs a learnable attention mechanism that computes attention weights over neighboring nodes, allowing the network to focus on more relevant neighbors during message passing. In this work, we leverage the Graph Attention Network (GAT) variant GATv2, which offers the ability to dynamically compute attention weights as opposed to the static computation in the standard GAT architecture.

\subsection{Assembly Graph Representation}
Product assemblies are often composed of many parts that are physically connected together and organized in a hierarchy. By viewing the parts as nodes and the connections as edges, it is possible to represent assemblies as graphs. 

Prior work has used graph edges to represent different types of connections between parts: joints~\cite{willis2022joinable}, which define the relationship between two parts whose relative pose and degree-of-freedom are constrained, and contacts~\cite{bian2024hg}, which define the relationship between two parts whose faces are touching. While joints have to be manually defined by the designer through the use of a CAD program (thus making them more error-prone and likely not to be included to save time during the design process), contacts can be computed from the completed CAD model by measuring the distance between all faces within a specific tolerance (resulting in a more reliable edge type). However, the contact's reliance on a pre-defined tolerance can lead to inaccurate or disconnected assembly graphs, which do not properly represent the connections between parts in assemblies. 

Given the large data required to train machine learning models, large collections of CAD assemblies have been publicly released, including the ABC dataset~\cite{Koch_2019_CVPR}, the Fusion 360 Gallery Assembly dataset~\cite{willis2020fusion}, and the Mechanical Components Benchmark~\cite{sangpil2020large}, among others. These datasets have fueled the development of data-driven design methods, as they provide a substantial source of complete assembly designs that include data such as part names, assembly names, part 3D geometry, part materials, and connections between parts. However, these datasets either omit, in the case of ABC, or provide erroneous interface information, in the case of the Fusion 360 Gallery dataset. 

In our work, we recompute the interface information for the Fusion 360 Gallery dataset and make it available as a simple overwrite operation. 
We create interface-informed assembly graphs where edges are embedded with the respective local part interface information and experiment with different methods for embedding this interface information.
\begin{figure*}[]
	\centering
	\includegraphics[width=0.95\linewidth,trim={0in 1.25in 0in 1.25in},clip]{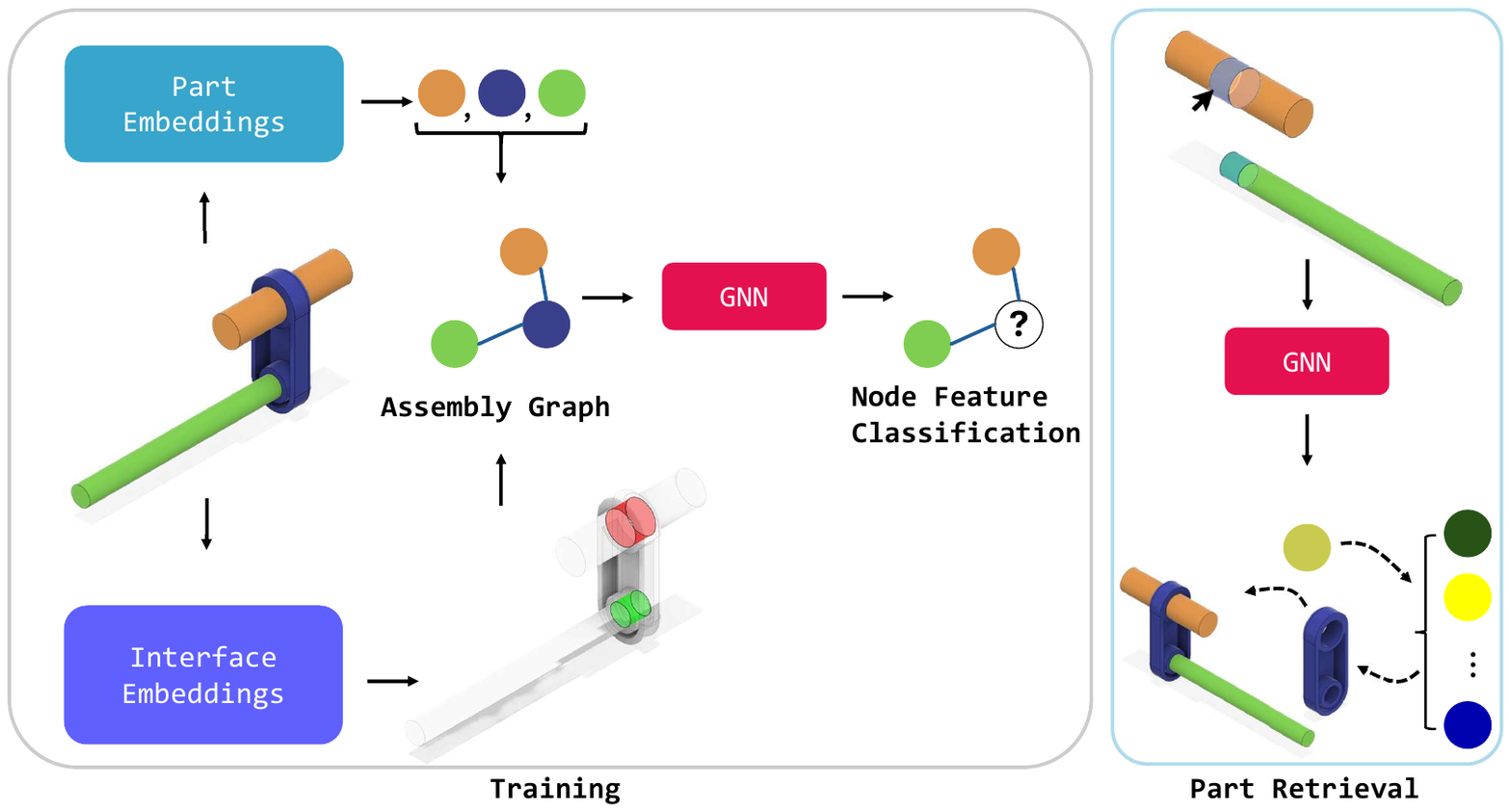}
	\Description{Overview of our proposed approach.}
	\caption{Overview of our proposed approach. During training, the input to our GNN is a batch of interface-informed assembly graphs. Each assembly graph is augmented with node attributes (part embeddings) and edge attributes (interface embeddings). The network is trained in a multi-class classification setting. During inference, we leverage the trained network for part retrieval by providing an incomplete assembly and predicting its cluster label.}
	\label{fig:overview}
\end{figure*}

\subsection{Missing Part Prediction}
A common approach in the literature is to train machine learning models to predict masked features, thus aiding designers by creating recommendation models. For example, Meltzer \textit{et al.} fine-tune a Large Language Model (LLM) on a corpus of part names organized in assemblies, derived from the ABC dataset, to predict a masked part from the list of part names in an assembly~\cite{meltzer2024s}. Bian \textit{et al.} develop a hierarchical GNN model to predict the material of parts in assemblies, given the part name, geometry, and assembly connections found in the Fusion 360 Gallery dataset~\cite{bian2022material,bian2024hg}. Gajek \textit{et al.} leverage a custom dataset of off-the-shelf components organized in assemblies to train a GNN for a missing part prediction task~\cite{gajek2022recommendation}. Their work excludes 3D geometry information and solely uses unique component identifiers as design features. Moreover, their largest dataset comprises only 3,099 unique parts, and the part recommendation does not include information about which other parts in the assembly it should be connected to. Another line of research has looked into 3D reconstruction using CAD models prior~\cite{gumeli2022roca, gao2024diffcad}, and they primarily focus on probabilistic approaches for CAD model retrieval and alignment from RGB images and do not consider assemblies. A concurrent work~\cite{mandelli2022cad} is similar in terms of the CAD model classification problem; however, it operates only on nodes and utilizes Graph Convolutional layers. 

Our work builds on existing literature and expands on the missing part prediction task in several ways. We consider 3D geometry information as a design feature, which allows us to generalize our method to bespoke parts and not just a limited set of off-the-shelf components. 
 
We consider the connection of the missing part to the other parts in the assembly, which is more aligned with how a designer might use this model. We also introduce a novel GNN architecture and assembly graph representation that does not rely on joints, thus creating the opportunity to leverage other datasets in the future that might lack this metadata.

\begin{figure*}[]
    \centering
    \begin{subfigure}[b]{0.3\textwidth}
        \includegraphics[width=\textwidth]{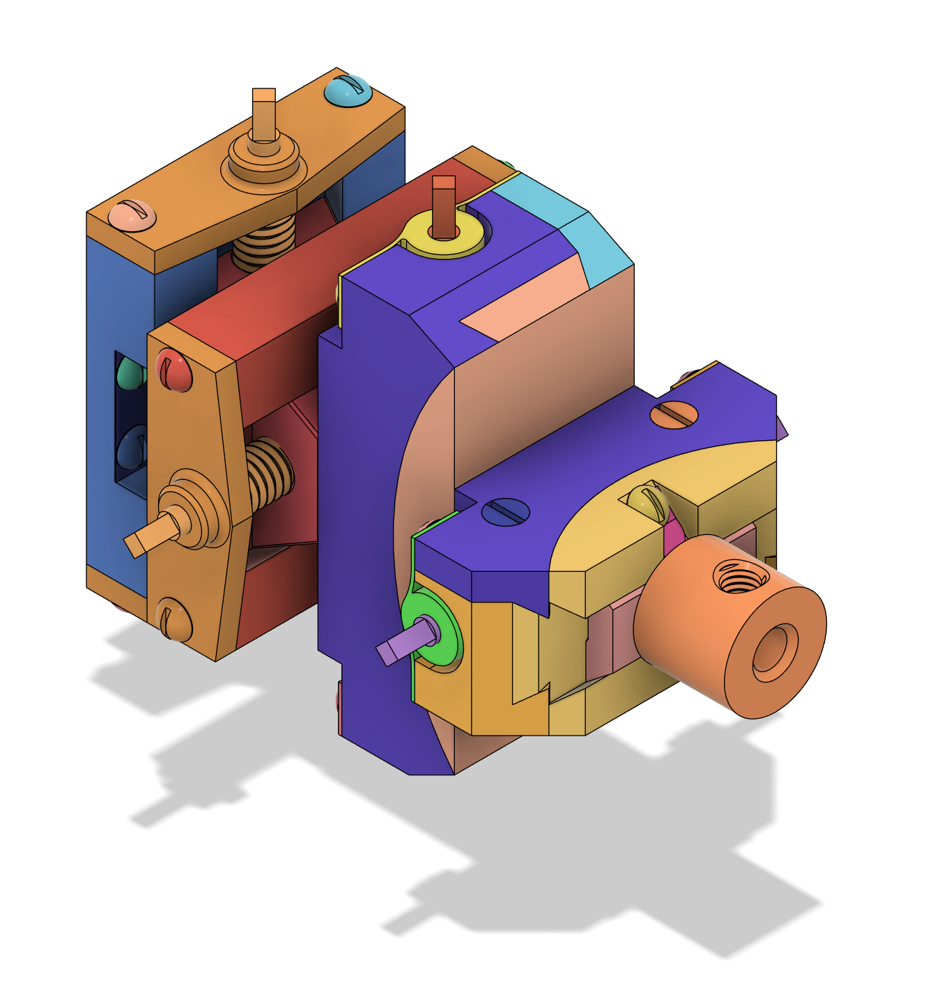}
        \caption{Assembly}
        \label{fig:regenerated-contacts-1}
    \end{subfigure}
    ~
    \begin{subfigure}[b]{0.3\textwidth}
        \includegraphics[width=\textwidth]{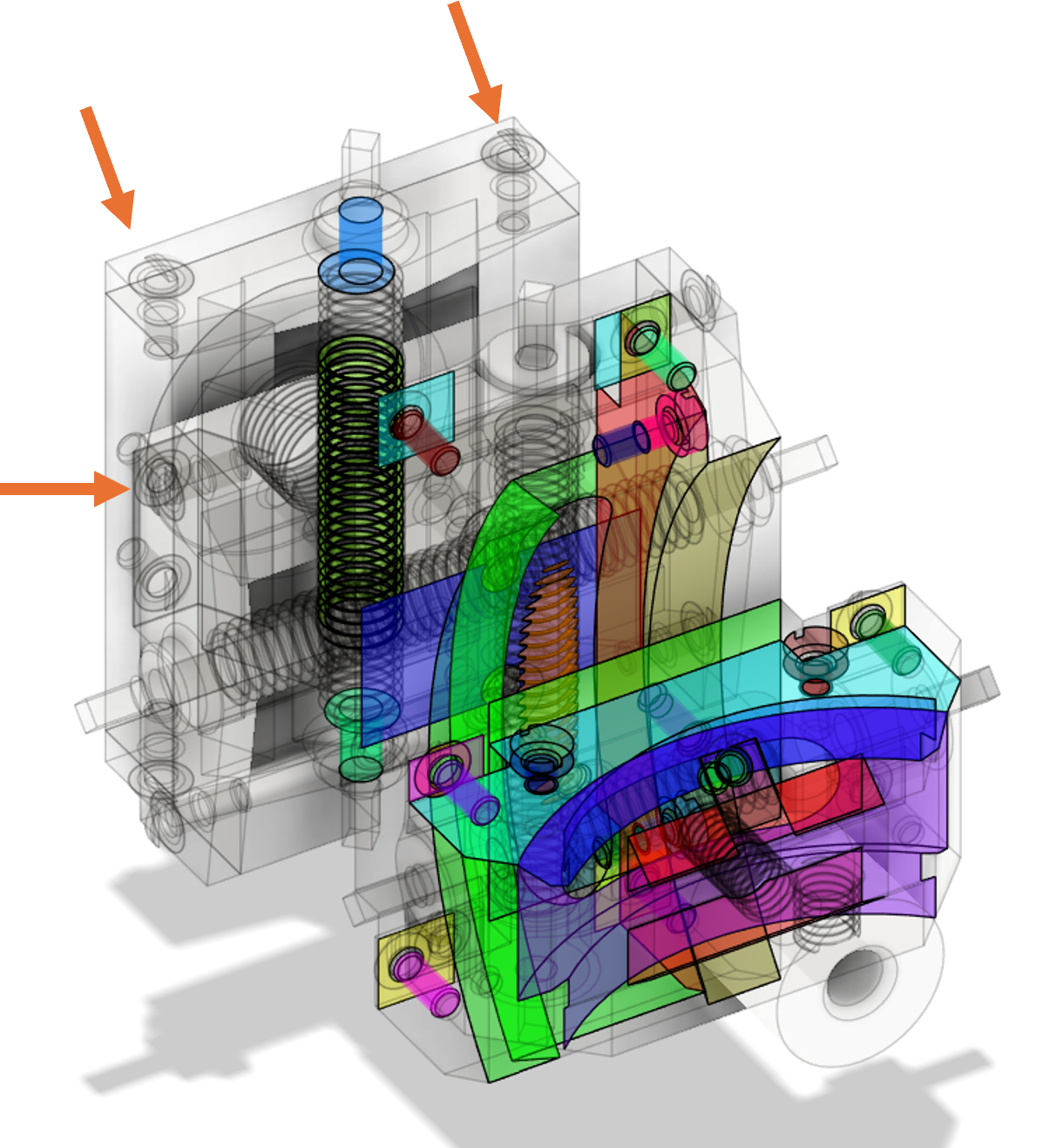}
        \caption{Fusion 360 interfaces}
        \label{fig:regenerated-contacts-2}
    \end{subfigure}
    ~
    \begin{subfigure}[b]{0.3\textwidth}
        \includegraphics[width=\textwidth]{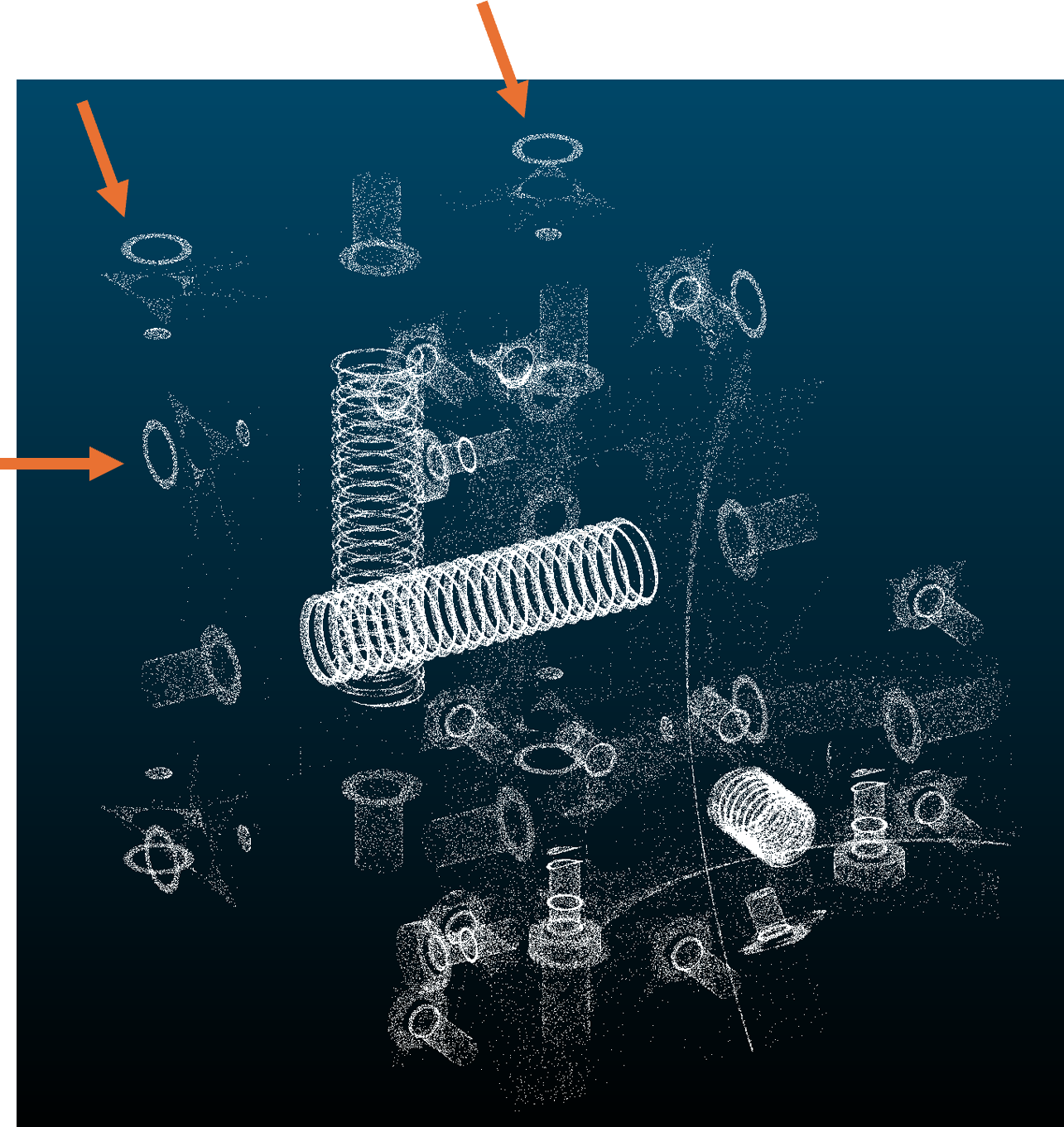}
        \caption{Our recomputed interfaces}
        \label{fig:regenerated-contacts-3}
    \end{subfigure}
    \caption{An example of an assembly with many parts (\ref{fig:regenerated-contacts-1}), showcasing missing interfaces in Fusion 360 between the screws and the flat plates (\ref{fig:regenerated-contacts-2}), and the pointcloud interfaces we recompute (\ref{fig:regenerated-contacts-3}). The orange arrows indicate three examples of contacts that our script correctly generates.}
        \label{fig:regenerated-contacts}
\end{figure*}

\section{Methodology}\label{sec:methods}
Our proposed approach aims to explore the use of valuable interface information between parts of an assembly and answer the question, ``Are interfaces useful for downstream tasks?" We showcase the overall pipeline in \figref{fig:overview}. In this section, we present the components of our approach. First, we introduce our dataset (\secref{sec:dataset}) and its processing scheme. Next, we describe the part prediction task (\secref{sec:partprediction}), ablation studies (\secref{sec:ablation}), and finally, our model architecture (\secref{sec:model}).

\subsection{Interface-Aware Dataset}{\label{sec:dataset}}
We build on the Fusion 360 Gallery Assembly dataset, as it provides interface information for each assembly in addition to hierarchically representing them as Assembly Graphs~\cite{willis2022joinable}. Each assembly's metadata consists of part names, part properties (center of mass, area, volume, density, mass, material), interface information (type of interface, bounding box, face index), and joints. Each part in the assembly utilizes a user-defined occurrence hierarchy. The main Fusion-360-specific dependency in our pipeline is the assembly metadata format, not the learning architecture itself. More generally, our method requires part-level geometry, assembly structure, and a way to derive or read inter-part relations. Porting the method to another CAD source would therefore mainly require an adapter for that dataset's hierarchy, geometry references, and assembly relations. If part geometry and assembly structure are already available, the main remaining cost is interface computation at scale; however, our code for this stage can be reused directly on STEP files.

The Fusion 360 Gallery Assembly dataset consists of \texttt{8,251} assemblies and a total of \texttt{154,468} separate parts. We found that existing interfaces were not accurate and would often either be completely omitted for the whole assembly, as shown in \figref{fig:regenerated-contacts}, or were incomplete. To address this limitation, we recomputed interface information for all assemblies in the entire dataset using geometric face-body intersections to accurately identify contact regions between components. More specifically, we use native Boolean operations from the Open CASCADE Technology (OCCT) B-Rep kernel. For each pair of bodies whose bounding boxes overlap, we compute the geometric intersection of each face of one body with the neighboring solid using \texttt{BRepAlgoAPI\_Common(face, solid)}. This face-to-solid Boolean common operation returns the precise B-Rep contact region between components. During this recomputation process, we also sample point clouds on the interface volumes and up-sample them if they do not have a minimum of \texttt{2,048} points. Importantly, this sampling is performed only after the exact interface geometry has already been computed, and is used solely to generate the neural network input representation. From the resulting intersection shape, we compute interface area and volume analytically using OCCT's built-in geometric property routines, and we additionally retain the indices of the contacting faces. The recomputation process was further enhanced by augmenting the interface data with the \texttt{interface area, interface volume}, and \texttt{indices of faces in contact}. To handle the large-scale dataset, the pipeline was parallelized, enabling the processing of multiple assemblies simultaneously. Furthermore, we perform filtering to keep assemblies with \texttt{num bodies} $\geq$ 4. This filtration resulted in a total of \texttt{6,039} valid assemblies.

\subsubsection{Assembly Graph Representation}
The input to our model is an interface-augmented assembly graph with node and edge attributes. For each assembly, we utilize the interface metadata to identify bodies in contact and use this information to generate an assembly graph, where nodes represent each body and edges represent the contacts between them. Mathematically, we denote a set of assembly graphs as $G = \{G_i\}_{i=1}^N$, with individual graphs represented as $G_i = (\mathcal{V}, \mathcal{E}, \mathcal{M}, \mathcal{Y})$ where $\mathcal{V} = \{v_1, v_2, ..., v_{|\mathcal{V}|}\}$ is the set of nodes, each with an associated attribute vector $\mathbf{x}_j \in \mathbb{R}^{d_v}$ describing its properties. The set of edges is denoted by $\mathcal{E} \subseteq \mathcal{V} \times \mathcal{V}$, where an edge $(v_j, v_k) \in \mathcal{E}$ exists if bodies $j$ and $k$ are in contact. Each edge also has an attribute $\mathbf{e}_{jk} \in \mathbb{R}^{d_e}$ that describes the interface. $\mathcal{M}$ represents a binary mask applied to the nodes for our pre-training task, and $\mathcal{Y}$ contains the ground-truth node labels used for downstream tasks or evaluation.

\subsubsection{Node Embeddings}
To equip the nodes of our assembly graph with meaningful features, we generate embeddings for each constituent part. This is achieved using a pretrained PointMAE encoder~\cite{10.1007/978-3-031-20086-1_35}, which was trained on the ShapeNet dataset \cite{chang2015shapenet}. For each part, represented by a node $v_b$, we first sample a point cloud $P_b = \{p_j\}_{j=1}^K$ from its tessellated surface, with the number of points $K = \texttt{2,048}$ and each point $p_j \in \mathbb{R}^3$. This fixed sampling budget is a practical approximation chosen for computational consistency across the dataset; we do not claim it is uniformly sufficient for all CAD geometries. The PointMAE model operates by partitioning the input point cloud $P_b$ into irregular point patches, of which 60\% are randomly masked. A standard Transformer-based autoencoder then learns high-level latent features from the remaining unmasked patches. We can represent this encoding process as a function, $f_{PointMAE}$, which maps the input point cloud to an embedding. The generation of the node attribute $x_b$ for node $v_b$ is thus given by $x_b = f_{\text{PointMAE}}(P_b)$. The resulting embedding $x_b$ is a \texttt{384}-dimensional vector.

In addition to PointMAE, we also generated node embeddings using a pretrained DGCNN~\cite{wang2019dynamic} encoder. This model was pretrained on the ModelNet40 dataset and produces a \texttt{256}-dimensional vector. Unlike PointMAE, which learns global geometric priors through masked reconstruction, DGCNN constructs a dynamic k-nearest neighbor graph at each layer and applies the EdgeConv operator to aggregate the local geometric neighborhoods. Including DGCNN allows us to evaluate two different encoders in their ability to accurately capture geometric features.

To introduce a categorical notion of geometric similarity, we perform unsupervised clustering over all node embeddings across the dataset. Specifically, we apply the \texttt{k-means} algorithm with $k=\texttt{500}$ clusters, where each cluster corresponds to a distinct geometric category learned purely from the latent space. We pick 500 clusters after a qualitative evaluation of the data and the duplicates found therein, attempting to keep the number of classes as high as possible while balancing task complexity and real-world applicability. This clustering process groups parts with similar morphological characteristics, such as fasteners, brackets, or housings, without relying on any manual annotation. The resulting cluster assignments are used as categorical node labels in $\mathcal{Y}$, serving as the ground-truth labels for our pre-training objective and subsequent node-level evaluation tasks. We emphasize that this choice defines the benchmark label space and is not intended as an intrinsic or canonical partition of mechanical part geometry.

\subsubsection{Edge Embeddings}
To encode the interface information into our assembly graph, we generate embeddings for each edge, adopting a methodology similar to our node embeddings. For any given edge $e_{ij} \in \mathcal{E}$, which represents the contact interface between the parts corresponding to nodes $v_i$ and $v_j$, we consider its geometric representation, $S_{ij}$, derived from our face-body intersection computations. This representation is obtained from the exact OCCT Boolean intersection geometry, not from point-based proximity tests. From this volume, we sample a point cloud $P_{ij}=\{p_k\}_{k=1}^K$, ensuring a consistent size of $K = \texttt{2,048}$ points, where each point $p_k \in \mathbb{R}^3$. To generate this point cloud, we tessellate the B-Rep intersection shape using OCCT's mesher and then sample points on the tessellated interface surface. This tessellation-based sampling is used only to construct a learned geometric embedding of the interface; it is not used to determine whether contact exists, nor to compute interface area or volume. This point cloud, which captures the geometry of the interface, is then processed by the same pretrained PointMAE encoder $f_{PointMAE}$, resulting in a 384-dimension vector $e_{ij}$, such that $e_{ij}=f_{PointMAE}(P_{ij})$.

\subsection{Part Prediction}{\label{sec:partprediction}}
To test the usefulness of this interface data, we leverage a part prediction task for incomplete assemblies. More specifically, given an existing interface-embedded assembly graph with one of the parts held out, the model must predict the cluster label of the missing part from a known vocabulary. We formulate this as a node classification problem on graphs. Mathematically, let $G=(\mathcal{V}, \mathcal{E})$ be an assembly graph with the node features $X=\{ x_b | v_b \in \mathcal{V}\}$ and edge features $\mathcal{E}=\{ e_{ij} | e_{ij} \in \mathcal{E}\}$. During data processing, we randomly select a target node $v_m \in \mathcal{V}$ to mask. We construct a corrupted graph $G'$ by replacing the feature vector of the masked node with a zero vector. All other node and edge embeddings remain unchanged.

A GNN, denoted by $f_\text{GNN}$, is tasked with learning the identity of the masked node by aggregating information from its local neighborhood. The GNN takes the corrupted graph $G'$ as input and produces a set of context-aware node embeddings, $\{\mathbf{h}_v\}_{v \in \mathcal{V}}$:

\begin{equation}
\{\mathbf{h}_v\}_{v \in \mathcal{V}}=f_\text{GNN}(G', X_{masked}, \mathcal{E}_{attr})
\end{equation}

The final embedding of the masked node, $\mathbf{h}_m$, is passed through a classifier head, $g_\text{cls}$, to predict the node's original class label, $y_m \in \mathcal{Y}$.

\begin{equation}
\hat{y}_m=g_{cls}(\mathbf{h}_m)
\end{equation}

\subsubsection{Evaluation metric}
We evaluate our GNN and baseline methods using a comprehensive set of metrics for multi-class classification. As the task involves a large number of classes for part-retrieval, we report Top-K accuracy for K values of \{1, 3, 5, 10, and 50\}. To provide a complete performance picture and account for a potential class imbalance, we also report the weighted scores for Precision, Recall, and F1-score.

\subsubsection{Baseline models}
To establish a benchmark, we compare our GNN against several baselines that deliberately ignore the graph's structure and edge features. The following methods are used:

\begin{itemize}
    \item \textbf{Random Guess (RG):} A lower-bound based baseline that predicts the class uniformly at random from all classes.
    \item \textbf{Majority Class (MC):} A heuristic that always predicts the single most frequent class observed in the training data. This tests whether the model learns more than just the a priori distribution of parts.
    \item \textbf{k-Nearest Neighbors (k-NN):} A non-parametric model that classifies a masked node by taking a majority vote of its neighbors. To represent the local context without using the graph structure, the node attribute for the masked part is the mean of the node attributes of all its visible neighbors. This baseline tests if simple feature similarity of the local context is sufficient for part retrieval.
    \item \textbf{Logistic Regression (LogReg):} A linear model is trained on the mean embedding of the visible neighbor nodes to predict the masked node's class. This baseline evaluates the effectiveness of a simple linear mapping on the aggregated local features.
\end{itemize}

\begin{figure}[]
    \centering
    \includegraphics[width=0.99\linewidth,clip,trim={1.25in 1.75in 1.25in 1.75in}]{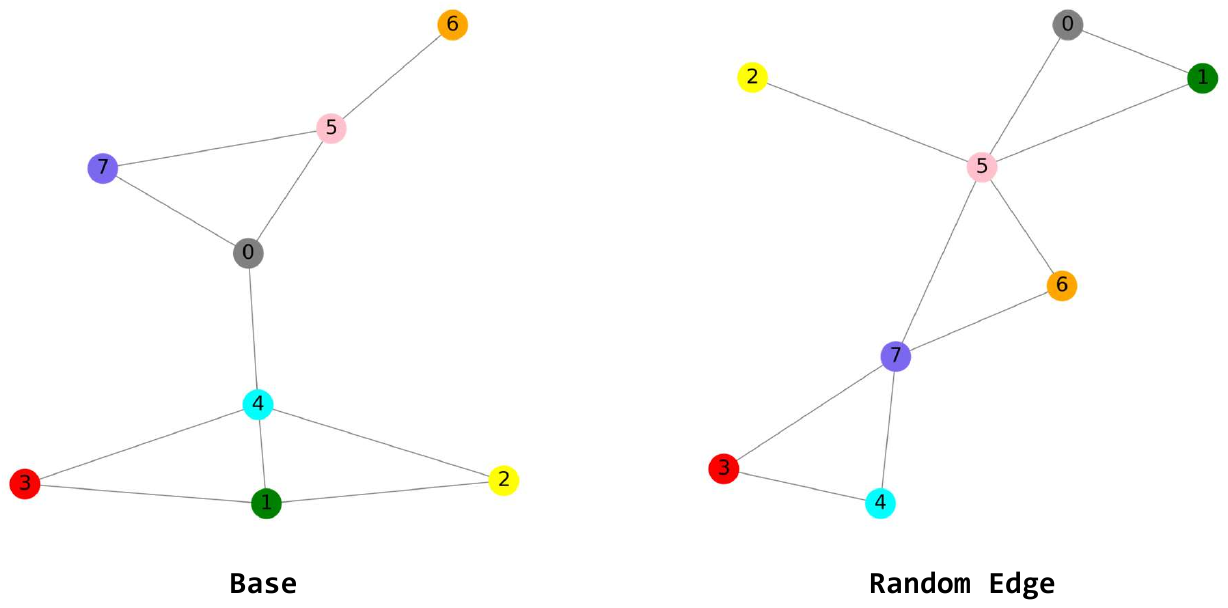}
    \caption{For each interface-aware Assembly Graph (Base), we generate a variant where the edge order is randomized.}
    \label{fig:ablation_re}
\end{figure}

\begin{figure}[]
    \centering
    \includegraphics[width=0.99\linewidth,clip,trim={1.0in 1.0in 1.0in 1.0in}]{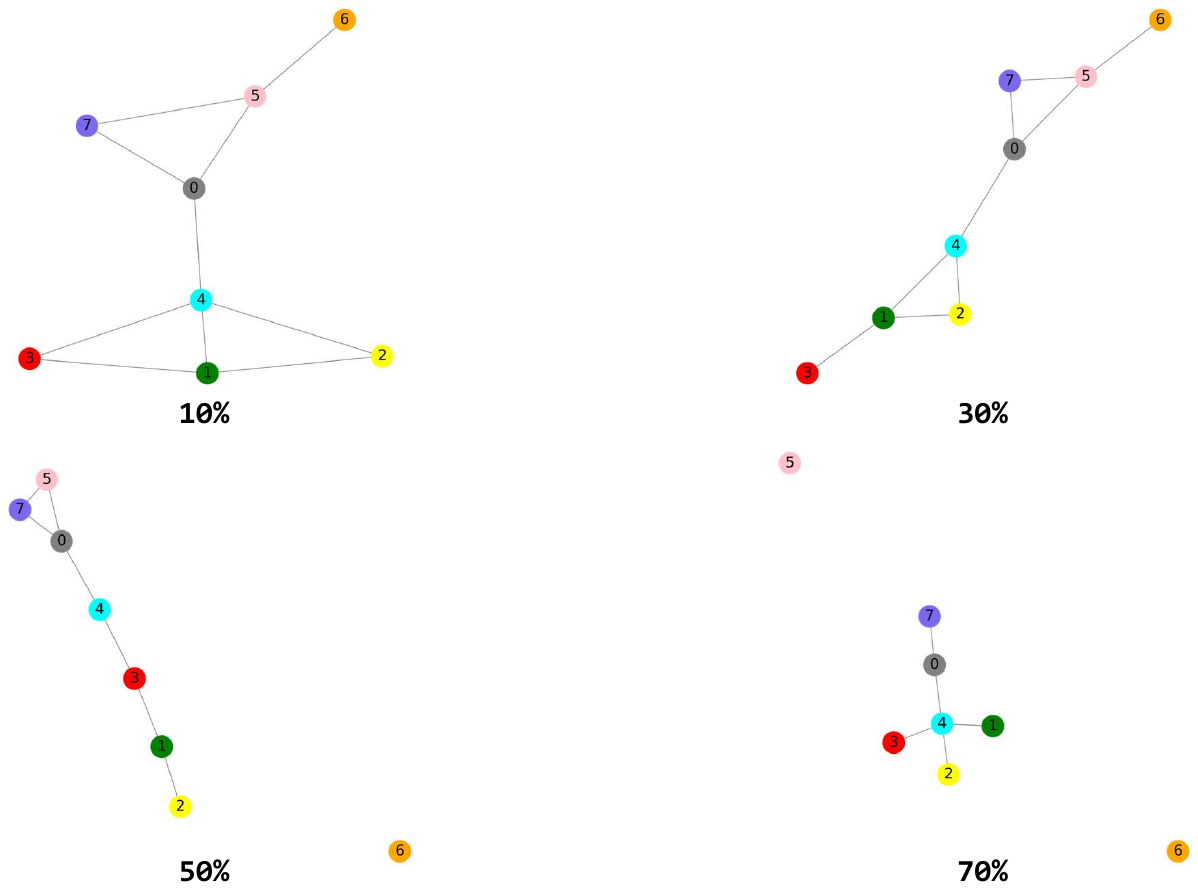}
    \caption{For each interface-aware Assembly Graph (Base), we also generate partially connected variants where the probability of edge removal is set to 10\%, 30\%, 50\%, and 70\%.}
    \label{fig:ablation_pc}
\end{figure}

\begin{table*}[t]
  \centering
  \setlength{\tabcolsep}{3pt}
  % \begin{tabular}{|c|c|c|c|c|c|}
  \begin{tabular}{@{}c@{\hspace{3pt}}cccccccc@{}} 
    \toprule
    Type & Top-1 & Top-3 & Top-5 & Top-10  & Top-50 & F1 & Precision & Recall \\
    \midrule
    RG              & 0.23 \(\pm\) 0.09 & 0.5 \(\pm\) 0.13  &  0.99 \(\pm\) 0.13  & 2.15 \(\pm\) 0.48 & 10.75 \(\pm\) 1.00 & 0.24 \(\pm\) 0.1 & 0.34 \(\pm\) 0.2 & 0.23 \(\pm\) 0.08\\
    MC    & 0.59 \(\pm\) 0.00 &      -            & -                   &        -           &      -              & $6 \times 10^{-5}$ \(\pm\) 0.0& $3 \times 10^{-5}$ \(\pm\) 0.0 & 0.59 \(\pm\) 0.0\\
    LogReg  & 5.99 \(\pm\) 0.0 & 13.46 \(\pm\) 0.0 & 16.89 \(\pm\) 0.0  & \textbf{25.93 \(\pm\) 0.0}   &  \textbf{50.98 \(\pm\) 0.0}  & 4.68 \(\pm\) 0.0& 4.67 \(\pm\) 0.0 & 5.99 \(\pm\) 0.0\\
    k-NN                &  3.93 \(\pm\) 0.0 & 7.23 \(\pm\) 0.0  & 12.08 \(\pm\) 0.0  & 13.36 \(\pm\) 0.0    & 21.61 \(\pm\) 0.0    &  3.65 \(\pm\) 0.0  & 4.26 \(\pm\) 0.0 & 3.93 \(\pm\) 0.0\\
    Ours-PointMAE   & \textbf{6.34 \(\pm\) 0.67} & \textbf{13.02 \(\pm\) 0.47} & \textbf{16.92 \(\pm\) 0.66} & 23.56 \(\pm\) 1.07 & 44.76 \(\pm\) 0.97 & \textbf{6.01 \(\pm\) 0.71}& \textbf{7.8 \(\pm\) 1.26}& \textbf{6.34 \(\pm\) 0.67}\\
    \bottomrule
  \end{tabular}
  \caption{Quantitative performance of our proposed method using PointMAE as a point cloud encoder. We compare against the following baselines - random (RG), majority class (MC), logistic regression (LogReg), and K-nearest neighbors (k-NN).}
  \label{tab:quantitative_pointmae}
\end{table*}
\vspace{-0.5em}

\begin{table*}[!h]
  \centering
  \setlength{\tabcolsep}{3pt}
  % \begin{tabular}{|c|c|c|c|c|c|}
  \begin{tabular}{@{}c@{\hspace{3pt}}cccccccc@{}} 
    \toprule
    Type & Top-1 & Top-3 & Top-5 & Top-10  & Top-50 & F1 & Precision & Recall \\
    \midrule
    RG              & 0.21 \(\pm\) 0.16 & 0.59 \(\pm\) 0.17 & 0.95 \(\pm\) 0.22 & 2.93 \(\pm\) 0.31&9.9 \(\pm\) 0.71  & 0.19 \(\pm\) 0.14 & 0.25 \(\pm\) 0.19 &0.21 \(\pm\) 0.16 \\
    MC    & 0.88 \(\pm\) 0.0 & - & - & - & - & 0.02 \(\pm\) 0.0 & $7 \times 10^{-5}$ \(\pm\) 0.0 & 0.88 \(\pm\) 0.0\\
    LogReg              & \textbf{4.72 \(\pm\) 0.0} & \textbf{11.78 \(\pm\) 0.0} & \textbf{15.71 \(\pm\) 0.0} & \textbf{22.59 \(\pm\) 0.0} & \textbf{45.78 \(\pm\) 0.00} & 2.81 \(\pm\) 0.0 & 2.44 \(\pm\) 0.0 & \textbf{4.72 \(\pm\) 0.0}\\
    k-NN                & 3.44 \(\pm\) 0.0 & 7.17 \(\pm\) 0.0 & 10.71 \(\pm\) 0.0 & 11.69 \(\pm\) 0.0& 17.1 \(\pm\) 0.0 & \textbf{3.03 \(\pm\) 0.0} & \textbf{4.68 \(\pm\) 0.0} & 3.44 \(\pm\) 0.0\\
    Ours-DGCNN               & 1.01 \(\pm\) 0.57 & 2.8 \(\pm\) 1.42 & 4.55 \(\pm\) 2.36 & 7.92 \(\pm\) 3.71 &  25.92 \(\pm\) 7.55 &  0.63 \(\pm\) 0.52&  0.74 \(\pm\) 0.67& 1.01 \(\pm\) 0.57\\
    \bottomrule
  \end{tabular}
  \caption{Quantitative performance of our proposed method using DGCNN as a point cloud encoder. We compare against the following baselines - random (RG), majority class (MC), logistic regression (LogReg), and K-nearest neighbors (k-NN).}
  \label{tab:quantitative_dgcnn}
\end{table*}

\subsection{Ablation Studies}
\label{sec:ablation}
To validate our architectural choices and understand the contributions of each key component, we conduct a series of ablation studies.

To address RQ 1, we test how two different data representation techniques, \textbf{Random Edge} (\figref{fig:ablation_re}) and \textbf{Partial Connectivity} (\figref{fig:ablation_pc}), affect learning. For the \textbf{Random Edge} case, the nodes and their features are preserved, but the edges are rewired. We generate a random spanning tree to ensure graph connectivity, and random edges are added to match the original edge count. This tests whether the model is truly leveraging the specific connections within the assembly or simply using an aggregated bag-of-parts representation. In the \textbf{Partial Connectivity} case, we systematically remove edges at varying probabilities (10\%, 30\%, 50\%, and 70\%) to understand the model's sensitivity to missing connections. For all cases, edge attributes are not considered, we only test using the presence or absence of the edge.

To address RQ 2, we evaluate two versions of our model's edge attributes---\textbf{No Edge Attributes} and \textbf{Random Edge Attributes}. For the former, we repurpose our model to not use any edge features, forcing the model to learn purely from node attributes. In the latter case, to verify the semantic viability of our interface embeddings, we replace them with randomly generated embeddings of the same dimension (384).

To address RQ 3, we directly compare GATv2's performance against the original GAT formulation. We train an identical model in which the GATv2Conv layers are replaced with standard GATConv layers and utilize the same hyperparameters as our GATv2 model.

% \vspace{-0.5em}

\begin{figure*}[t!]
    \centering
    \includegraphics[width=0.99\linewidth,clip,trim={0.0in 1.0in 0.0in 0.0in}]{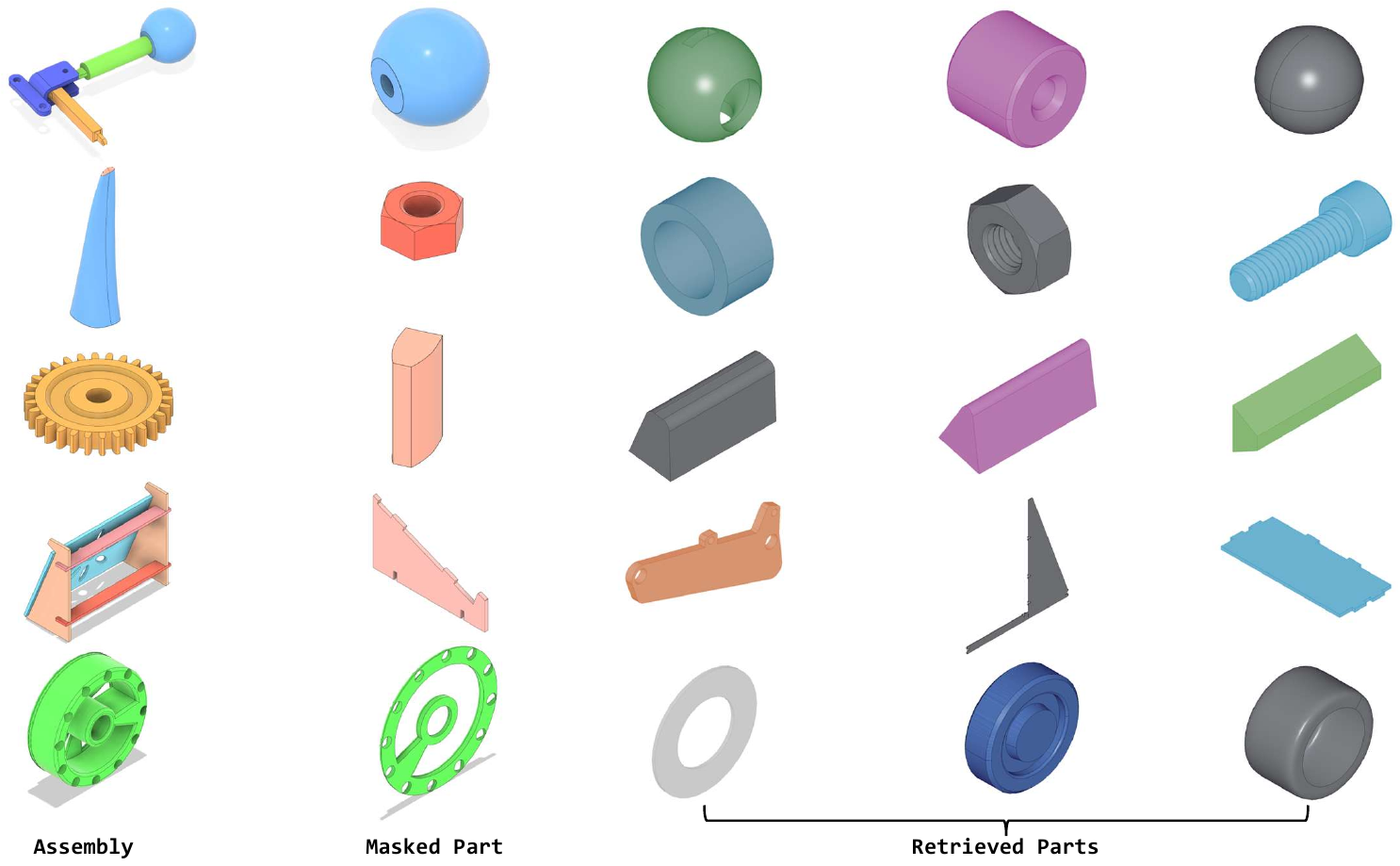}
    \caption{On each row, given an assembly and its masked out part, we showcase the successful retrieval of parts.}
    \label{fig:retrieval_success}
\end{figure*}

\begin{figure*}[t!]
    \centering
    \includegraphics[width=0.99\linewidth,clip,trim={0.0in 1.0in 0.0in 0.0in}]{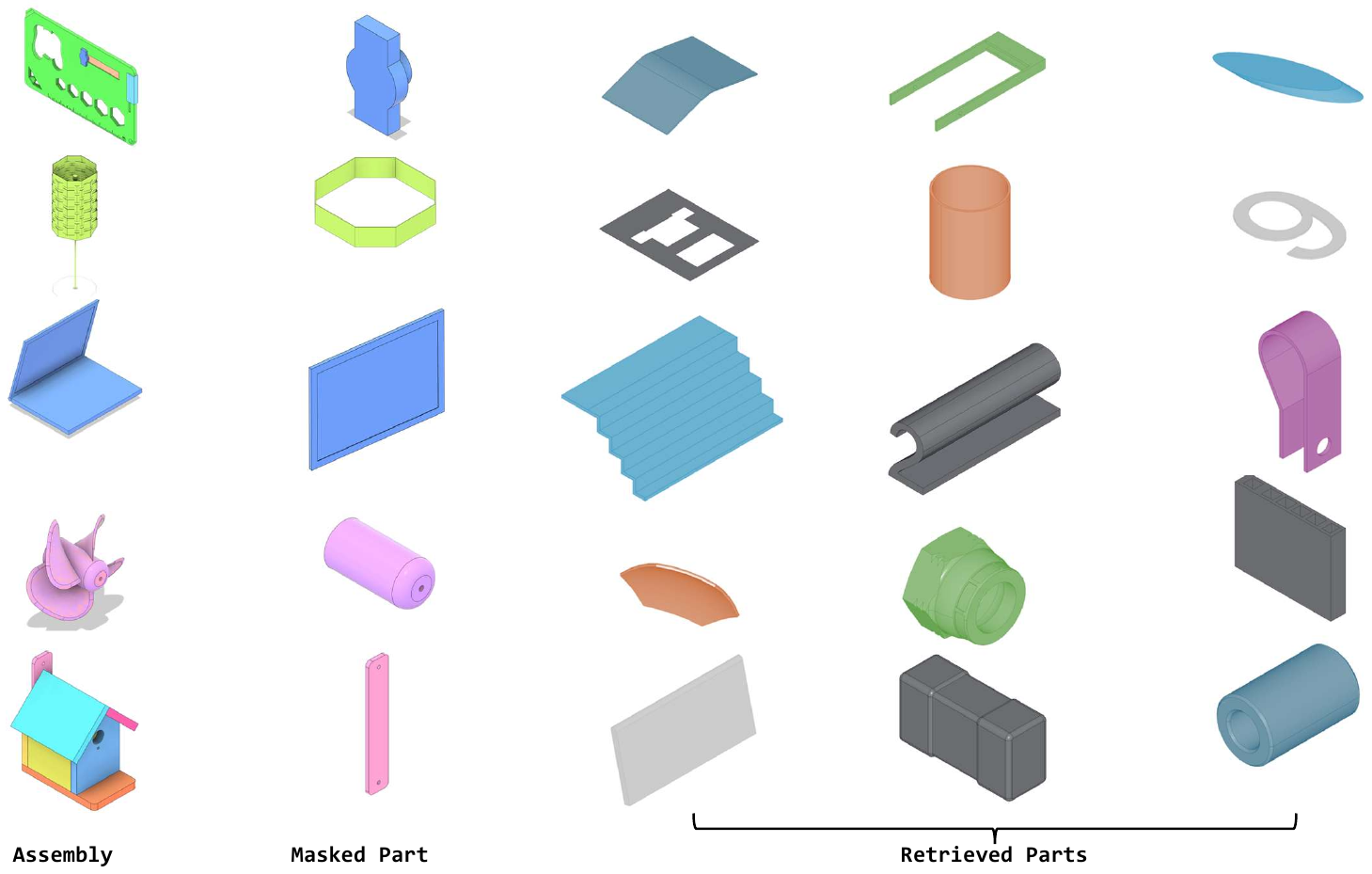}
    \caption{On each row, given an assembly and its masked-out part, we showcase failed retrieval of parts.}
    \label{fig:retrieval_bad}
\end{figure*}

\subsection{Implementation}{\label{sec:model}}
The core of our architecture is a Graph Attention Network (GAT) that leverages the more expressive \texttt{GATv2Conv} layer~\cite{brody2022attentivegraphattentionnetworks}. The GAT architecture computes the node representations by attending to their neighbors, following a self-attention strategy. The attention mechanism allows the model to dynamically assign different importance weights to different nodes within a neighborhood. The key advantage of \texttt{GATv2} over the original \texttt{GAT} is its use of dynamic attention. Instead of computing the attention weights based on a static function of the source and target node features, \texttt{GATv2} computes attention weights as a function of the source node, target node, and the query itself. This allows the model to compute more expressive, context-dependent attention scores for each node pair. We also incorporate dropout on the attention weights within each GAT layer and on the node features between layers to regularize the model.

We train the model using cross-entropy loss. To counter the significant class imbalance inherent in the part frequency distribution, we apply class weighting. The weight for each class is computed as the inverse of its frequency in the training set. We make use of the Adam optimizer~\cite{kingma_2014_iclr} with weight decay, a learning rate of \texttt{0.000068}, along with a cosine annealing schedule.

\section{Results and Discussion}\label{sec:resultsdiscussion}

\subsection{Quantitative Results}

We showcase the performance of our model using PointMAE (\tabref{tab:quantitative_pointmae}) and DGCNN (\tabref{tab:quantitative_dgcnn}) against our baselines. 

When using Point-MAE, we observe modest absolute accuracy---as expected for a 500-way fine-grained retrieval task---but also consistent improvements over the baselines. These gains suggest that the interface-aware assembly structure contributes additional signal beyond what can be captured from local mean-pooled features alone. Specifically, our approach achieves a Top-1 accuracy of \textbf{6.34 \(\pm\) 0.67}, improving upon the strongest baseline. Our F-1 score achieves a 28\% relative improvement over logistic regression. Notably, the precision-recall tradeoff favors precision (\textbf{7.8 \(\pm\) 1.26}), suggesting that when our model retrieves a part match, it is more likely to be correct.

When using DGCNN as a point cloud encoder, our model does not perform as well as the logistic regression model. Across all metrics, models trained on the DGCNN features perform worse than those trained on Point-MAE features. This underscores that encoder quality, particularly the mismatch between DGCNN's ModelNet40 pretraining and engineered CAD geometry, is currently a larger bottleneck than the graph architecture itself.

The standard deviations of our methods' performance indicate reasonable stability across different runs, though they are higher than those of deterministic baselines. The increasing accuracy from Top-1 to Top-50 suggests that, while exact matches remain challenging, the model successfully narrows down the search space to relevant part categories.

\begin{table*}[t!]
  \centering
  \setlength{\tabcolsep}{4pt}
  \begin{tabular}{@{}c@{\hspace{6pt}}ccccccccc@{}} 
  \toprule
    Type & Top-1 & Top-3 & Top-5 & Top-10 & Top-50 & F1 & Precision & Recall \\
    \midrule
    RE   &  5.48 \(\pm\) 0.64   &  10.72 \(\pm\) 0.69 & 14.21 \(\pm\) 0.89  &  20.76 \(\pm\) 1.02 & 40.61 \(\pm\) 1.48 & 5.03 \(\pm\) 0.57 & 6.15 \(\pm\) 0.81 & 5.48 \(\pm\) 0.64\\ 
    PC@10 &  4.83 \(\pm\) 0.85  & 10.07 \(\pm\) 1.16 &  13.72 \(\pm\) 1.12 & 19.56 \(\pm\) 1.24 & 38.11 \(\pm\) 2.53 &  4.7 \(\pm\) 0.83 & 6.02 \(\pm\) 1.14 & 4.83 \(\pm\) 0.85\\
    PC@30 & 3.64 \(\pm\) 0.53  &  7.83 \(\pm\) 0.82 & 10.4 \(\pm\) 0.88  & 15.02 \(\pm\) 1.32  & 32.58 \(\pm\) 1.03 & 3.54 \(\pm\) 0.63 & 4.92 \(\pm\) 1.05 & 3.64 \(\pm\) 0.53\\
    % FC  & -  & - & - & - & - & -& -& -\\
    PC@50 & 2.6 \(\pm\) 0.38  &  5.07 \(\pm\) 0.54 & 6.9 \(\pm\) 0.69  & 10.35 \(\pm\) 0.44  & 24 \(\pm\) 0.82 &  2.68 \(\pm\) 0.4 &  3.98 \(\pm\) 0.78 & 2.6 \(\pm\) 0.37\\
    PC@70 & 1.33 \(\pm\) 0.31 &  2.78 \(\pm\) 0.48 & 3.86 \(\pm\) 0.47 & 6.08 \(\pm\) 0.79  & 15.57 \(\pm\) 1.35 & 1.32 \(\pm\) 0.4 & 1.95 \(\pm\) 0.88 & 1.33 \(\pm\) 0.31\\
    \bottomrule
  \end{tabular}
  \caption{Comparison of our data representation techniques: Random Edge (RE) and Partial Connectivity (PC) at varying edge drop probabilities.}
  \label{tab:ablation_1}
\end{table*}

\begin{table*}[t!]
  \centering
  \setlength{\tabcolsep}{4pt}
  \begin{tabular}{@{}c@{\hspace{6pt}}ccccccccc@{}} 
  \toprule
    Type & Top-1 & Top-3 & Top-5 & Top-10 & Top-50 & F1 & Precision & Recall \\
    \midrule
    NEA   & 6 \(\pm\) 0.33 & 12.37\(\pm\) 0.89 & 16.46 \(\pm\) 1.0  & 23.63 \(\pm\) 1.20 & 44.47 \(\pm\) 1.2& 5.77 \(\pm\) 0.4 &7.44 \(\pm\) 0.6 & 6 \(\pm\) 0.33\\ 
    REA  &  5.84 \(\pm\) 0.58  & 12.13 \(\pm\) 1.03  & 15.73 \(\pm\) 1.33  & 22.68 \(\pm\) 1.34  & 44 \(\pm\) 1.31 & 5.57 \(\pm\) 0.48& 7.17 \(\pm\) 0.56& 5.84 \(\pm\) 0.58 \\
    Ours-PointMAE   & \textbf{6.34 \(\pm\) 0.67} & \textbf{13.02 \(\pm\) 0.47} & \textbf{16.92 \(\pm\) 0.66} & 23.56 \(\pm\) 1.07 & 44.76 \(\pm\) 0.97 & \textbf{6.01 \(\pm\) 0.71}& \textbf{7.8 \(\pm\) 1.26}& \textbf{6.34 \(\pm\) 0.67}\\
    \bottomrule
  \end{tabular}
  \caption{Comparison of our data augmentation techniques: No Edge Attribute (NEA) and Random Edge Attribute (RE).}
  \label{tab:ablation_2}
\end{table*}

\begin{table*}[t!]
  \centering
  \setlength{\tabcolsep}{6pt}
  \begin{tabular}{@{}c@{\hspace{6pt}}cccccccc@{}} 
    \toprule
    Type & Top-1 & Top-3 & Top-5 & Top-10 & Top-50 & F1 & Precision & Recall \\
    \midrule
    GAT   &  5.46 \(\pm\) 0.73 & 11.24 \(\pm\) 0.66 & 15.3 \(\pm\) 0.96  & 22.09 \(\pm\) 1.15  & \textbf{45.58 \(\pm\) 1.23} &  5.26 \(\pm\) 0.75 & 7.2 \(\pm\) 1.05 & 5.46 \(\pm\) 0.73\\
    GATv2  & \textbf{6.34 \(\pm\) 0.67} & \textbf{13.02 \(\pm\) 0.47} & \textbf{16.92 \(\pm\) 0.66} & \textbf{23.56 \(\pm\) 1.07} & 44.76 \(\pm\) 0.97 & \textbf{6.01 \(\pm\) 0.71}& \textbf{7.8 \(\pm\) 1.26}& \textbf{6.34 \(\pm\) 0.67}\\
    \bottomrule
  \end{tabular}
  \caption{Quantitative comparison of the performance of GATv2 against the standard GAT architecture.}
  \label{tab:ablation_3}
\end{table*}
% \vspace{-0.5em}
\subsection{Qualitative Results}

We showcase successful part-retrieval results in \figref{fig:retrieval_success}. The retrieved parts exhibit strong geometric and functional similarity to the masked components. For instance, in the mechanical assembly examples (first and fifth rows), when a cylindrical component or gear tooth is masked, the model successfully retrieves other cylindrical parts/teeth with compatible interface geometries. Similarly, for the structural assembly (fourth row), the model retrieves geometrically similar parts with similar-looking mounting interfaces. These qualitative examples suggest that the interface-aware representation supports context-sensitive retrieval behavior, even when quantitative performance is limited. We caution, however, that these examples should be interpreted as evidence of partially context-sensitive retrieval behavior rather than proof of full geometric or kinematic compatibility.

 We also show failure cases in \figref{fig:retrieval_bad}. Analysis of these outputs reveals our model's limitations. In the birdhouse assembly (fifth row), the model fails to distinguish between structurally different components, retrieving generic rectangular and cylindrical shapes. The pink structural component (third row) shows another failure mode where the model retrieves parts with vaguely similar aspect ratios but fundamentally different interface types.

\subsection{Ablation Results}

\tabref{tab:ablation_1} shows the results of the ablation experiment aimed at answering RQ 1.
We find that \textbf{Random Edge} rewiring causes a 13.6\% drop in Top-1 accuracy, suggesting that the model can partially compensate for incorrect connectivity through the node features. However, the \textbf{Partial Connectivity} results indicate a near-linear performance degradation with edge removal. Removing just 10\% of the edges causes a 23.8\% performance drop. The steep performance difference confirms that our interface-informed assembly graph representation is useful for successful part retrieval and that the correct computation of the interfaces is critical.

The results in \tabref{tab:ablation_2} show that removing edge attributes entirely results in a relative 5.4\% drop in Top-1 accuracy compared to Ours-PointMAE, while random edge attributes result in a 7.9\% relative drop. This suggests that the current pretrained encoders limit the utility of edge attributes; with CAD-specific pretraining or joint optimization, we expect these semantics to play a larger role.

In response to RQ3, looking at \tabref{tab:ablation_3}, we conclude that GATv2 provides a 16.1\% improvement over the standard GAT architecture. This validates that the dynamic attention mechanism of GATv2 is critical for interface-informed assembly graphs.

\section{Limitations and Future Work}
While our method demonstrates that interface information between parts does help networks learn to perform part prediction tasks on assembly graphs, we identify several avenues for future research. Our results highlight three key challenges: (i) Our model struggles with parts that require an understanding of functional roles beyond geometric similarity. (ii) Some failures occur when parts have similar overall geometry but incompatible interface details, suggesting that our interface embeddings may not capture fine-grained interface requirements. (iii) The model occasionally ignores the broader assembly context and retrieves parts that might fit geometrically but are semantically inappropriate. Accordingly, the current method should be viewed as a retrieval-oriented probe of assembly understanding rather than a complete assembly auto-completion or validation pipeline. A fully closed-loop system would require richer interface descriptors, explicit kinematic or constraint information, and downstream geometric verification of candidate parts. These limitations partly stem from our interface embeddings that do not capture joint kinematics (revolute/fixed/prismatic) and quantitative interface properties (area, normal force). Furthermore, our reliance on a pretrained point cloud encoder (PointMAE) that was trained on ShapeNet (furniture, household items, vehicles) introduces some noise into the learning process and consequently affects accurate retrieval.

An important factor underlying several of these issues is the scale of available data. Our current evaluation is constrained to $\sim6K$ assemblies, which limits diversity in geometric and functional patterns. We believe that testing on a significantly larger dataset--both in the number of assemblies and in the diversity of engineered part categories--will help resolve many of these challenges by enabling the network to better disentangle geometric, functional, and contextual cues. Larger-scale training may also reveal emergent structure in the learned embedding space, improving both clustering ability and retrieval robustness.

Future work will therefore focus on scaling the dataset, refining embeddings with kinematic and physical attributes, and aligning part clusters with standard engineering taxonomies such as PartNet to introduce semantic grounding. We also plan to analyze attention distributions within the network to identify functionally significant subassemblies, offering interpretability and design insights. Ultimately, we envision extending this retrieval-based framework to a generative setting, where the interface-aware embeddings of a masked node condition a model to synthesize new parts or assemblies, evolving the system into a true design-assist tool.

\section{Conclusions}
\label{sec:conclusions}
This paper presents a novel exploration of interfaces and their incorporation into assembly graphs for an example downstream task like part-retrieval using graph neural networks. Unlike prior methods that treat parts in isolation or rely solely on geometric similarity, we demonstrate that assembly structure, and more specifically, the interface relationships between parts, provide critical information for identifying compatible components. Our key contributions include the computation of geometric interface data for the Fusion 360 assembly dataset (which was incorrect in the original data release), along with an interface-aware assembly graph representation that combines learned point cloud features with graph attention mechanisms. This work establishes part retrieval as a valuable probe task for assembly understanding and demonstrates that graph-based representations are essential for modeling hierarchical assembly relationships. While current performance suggests the problem remains far from solved, our dataset and interface-aware approach provide a foundation for future systems that could enable automated design validation, assembly planning, and ultimately, power generative design approaches that understand the interplay of parts within functional assemblies.
% \vspace{-0.5em}

%%
%% The next two lines define the bibliography style to be used, and
%% the bibliography file.
\bibliographystyle{ACM-Reference-Format}
\bibliography{refs}

%%
%% If your work has an appendix, this is the place to put it.
% \appendix

\end{document}